\documentclass{article}

\PassOptionsToPackage{numbers, compress}{natbib}

\usepackage{caption}
\usepackage{subcaption}
\usepackage{fontawesome5}
\usepackage{graphicx}
\usepackage{rotating}
\usepackage{multirow}
\usepackage{float}

\usepackage[final]{neurips_2023}

\usepackage[utf8]{inputenc} 
\usepackage[T1]{fontenc}    
\usepackage{hyperref}       
\usepackage{url}            
\usepackage{booktabs}       
\usepackage{amsfonts}       
\usepackage{nicefrac}       
\usepackage{microtype}      
\usepackage{xcolor}         

\usepackage{soul}
\usepackage{amsmath}

\DeclareMathOperator*{\argmin}{arg\,min}

\newcommand{\hlc}[2][yellow]{{%
    \colorlet{foo}{#1}%
    \sethlcolor{foo}\hl{#2}}%
}

\title{"What's important here?": Opportunities and Challenges of Using LLMs in Retrieving Information from Web Interfaces}

\author{%
  Faria Huq, Jeffrey P. Bigham, Nikolas Martelaro\\
  School of Computer Science\\
  Carnegie Mellon University\\
  \texttt{\{fhuq, jbigham, nmartela\}@cs.cmu.edu} \\
}

\begin{document}

\maketitle

\begin{abstract}
Large language models (LLMs) that have been trained on a corpus that includes large amount of code exhibit a remarkable ability to understand HTML code \cite{gur2023understanding}. As web interfaces are primarily constructed using HTML, we design an in-depth study to see how 
LLMs can be used to retrieve and locate important elements for a user given query ({\em i.e.} task description) in a web interface. In contrast with prior works, which primarily focused on autonomous web navigation, we decompose the problem as an even atomic operation - \textit{Can LLMs identify the important information in the web page for a user given query?} This decomposition enables us to scrutinize the current capabilities of LLMs and uncover the opportunities and challenges they present. Our empirical experiments show that while LLMs exhibit a reasonable level of performance in retrieving important UI elements, there is still a substantial room for improvement. We hope our investigation will inspire follow-up works in overcoming the current challenges in this domain.
\end{abstract}

\section{Introduction}

Web assistants capable of understanding natural language commands and retrieving relevant information from web user interfaces (UI) could significantly enhance human efficiency. With the recent advancement of LLMs and their potential in understanding HTML, research attempts have predominantly focused on LLM-driven web assistants capable of autonomous navigation \cite{LLM4UI, zhou2023webarena, deng2023mind2web, gur2023understanding, shi2017world}. However, as demonstrated by \cite{zhou2023webarena} and \cite{deng2023mind2web}, these assistants can not yet exceed 15\% task completion accuracy for real-world websites. Given their limited performance, these autonomous agents 
are yet to be a practical solution for day-to-day usage. 

\begin{figure}
    \centering
    \includegraphics[width=0.8\textwidth,keepaspectratio]{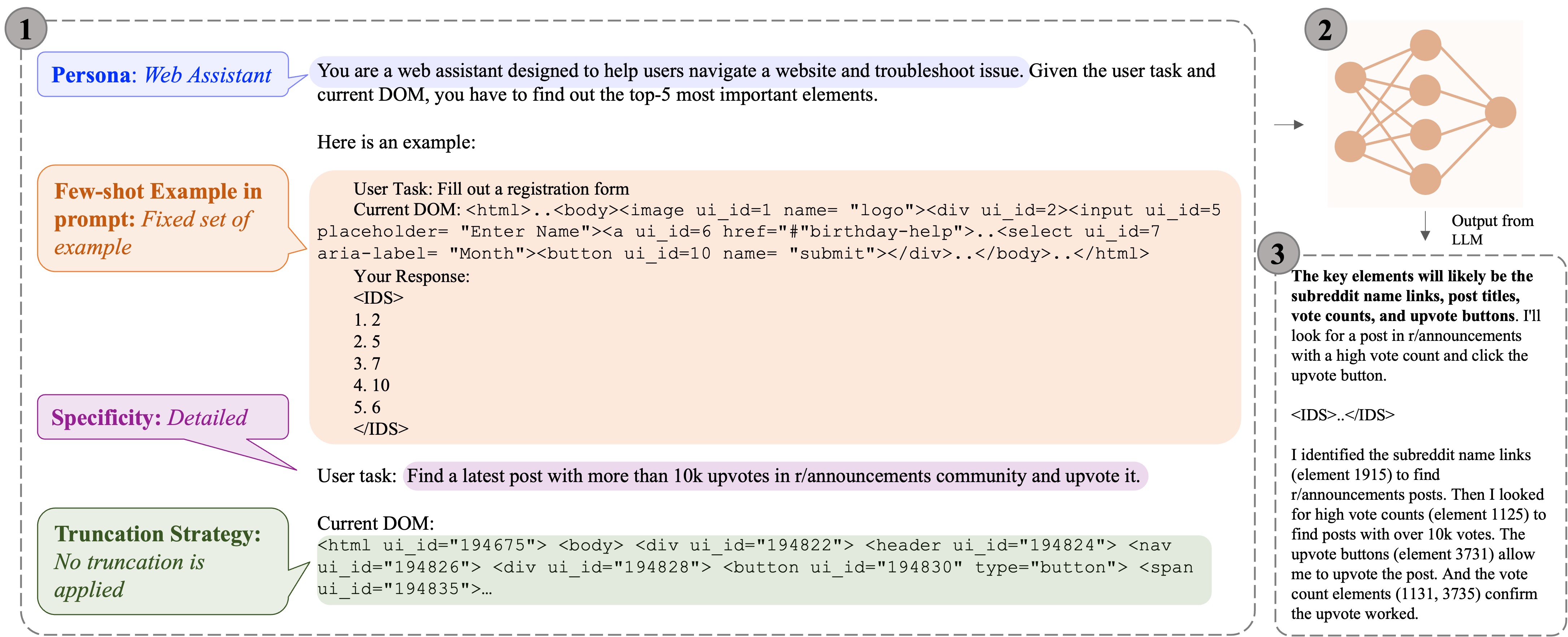}
    \caption{Overview of our experiment setup. (1) Input prompt is crafted based on four components, namely user role ({\em i.e: persona}), few-shot examples, specificity and HTML truncation strategy; (2) LLM (\texttt{Claude2} in this case) processes the given prompt; and (3) LLM generates the corresponding output.}
    \label{fig:sysmtem_img}
    \vspace{-10pt}
\end{figure}

To successfully complete a user given task that requires navigating through a series of web-pages, an autonomous agent must be able to perform a more fundamental operation -- locating and retrieving the most important UI elements from each page at each step of the corresponding task. 
Motivated by this insight, 
in this paper, we aim to explore LLMs' cabability to retrieve important and relevant elements in a webpage. 
%
%
We conducted a comprehensive investigation into four key components of an input prompt: 
1) \textit{Example selection for few-shot prompting}: How the selection of few shot examples impacts the performance;
2) \textit{Specificity of natural language command}: How the level of details in an input command impacts performance;
3) \textit{HTML truncation strategy}: How the strategy of HTML encoding impacts the performance; 
4) \textit{Specific role assumed by the LLM} (\textit{persona}): How the choice of role imitated by the LLM impacts the performance. (We refer readers to figure \ref{fig:sysmtem_img} to view an example.) 

We find that the performance of LLM depends on these components. Carefully prompting the few-shot examples can help LLMs to succeed as long as the input sequence size is reasonable. 
For instance, using 
few-shot examples that has semantically similar task descriptions, can boost the recall by 9.70\% in 1-shot prompting; however, it decreases the performance by 13.17\% in 2-shot prompting. We also show that an effective way to truncate the HTML document can alone lead to better performance with gains upto 11.54\%. We report critical limitations of LLMs such as hallucinating non-existing web elements and failure to follow input instruction. We conclude by providing future direction and possible solutions to the limitations we observed in our study.




\section{Related Works}

\subsection{LLM for UI}
LLM has recently gained popularity for many aspects of UI tasks. \cite{LLM4UI} shows an in-depth study on LLM for interacting with mobile UI --- ranging from task automation to screen summarization. \cite{wen2023empowering} performs an offline exploration and creates a transition graph, which is used to provide more contextual information to the LLM prompt. \cite{zhang2023look} introduces chain-of-action prompting that leverages previous action history and future action plans to decide the next action. These works primarily focus on Mobile UI -- which has a significantly smaller search space than Web UI. Compared to Mobile UI, Web UI is much more complex and has more elements on each page -- making it harder to process and incorporate with LLMs \cite{zhou2023webarena, deng2023mind2web}. Most of the early works on Web UI are based on synthetic frameworks, MiniWob \cite{shi2017world} and WebShop \cite{yao2023webshop}. \cite{gur2023understanding} proposed a fine-tuned T5 model on WebShop. However, these datasets are not well representative of real-world web activities. To capture the complexity of real world tasks, \cite{deng2023mind2web} and \cite{zhou2023webarena} introduce two realistic environments and datasets encompassing real-world tasks. 
However, to the best of our knowledge, there has been no in-depth analysis of the true capabilities of LLMs in the context of Web UI. 
Specifically, there is a gap in understanding how to effectively prompt LLMs for Web UI tasks, which prompting strategies yield favorable results and 
the underlying reasons behind their success or failure. 
In this study, we explore the potential of LLMs for Web UI information retrieval through evaluating the effectiveness of state-of-the-art prompting techniques.

\subsection{Prompting LLM for Instruction following }
LLMs are now largely used for many tasks related to instruction following \cite{han2023imagebindllm, liu2023visual, song2023llmplanner, fleming2023medalign, wang2023pandalm, li2023quantity, nakano2021webgpt}. Due to the large pool of works active in LLM, we only provide a high-level overview in this section. Especially for Vision-and-Language Navigation, LLM has been successful as a co-ordinator and/or planner \cite{Song_2023_ICCV, shah2022lmnav, zheng2022jarvis, zhou2023esc}. Some of the most recent works proposed new frameworks for holistic evaluation of LLMs \cite{wang2023boosting, shridhar2023screws, fernando2023promptbreeder, dhuliawala2023chainofverification, sumers2023cognitive}. However, these frameworks are mostly for analytic problem-solving for math and/or reasoning and may not be a proper reflection of the unique challenges encountered in tasks related to user interfaces. This study specifically focuses on exploiting some of these prompt techniques for Web UI and highlighting their feasibility in this domain.

\section{Experiments}
\label{gen_inst}

In our experiments, we use the \texttt{Claude2} model \footnote{For each experiment, we use a fixed value of temperature $ =~1$. We use a high temperature to encourage exploration\cite{zhou2023webarena}.} by Anthropic~\cite{Anthropic_2023}. 
\texttt{Claude2} has a context length of 100k tokens, which is the largest among all the LLMs to date. Having a large context window is especially beneficial for Web UI analysis considering the potentially thousands of elements on a webpage \cite{deng2023mind2web, gur2023understanding}.

\subsection{Problem Formulation}
We define each example as a set ``$\{\{w,q\},e\}$'', where $w$, $q$, and $e$ represent \texttt{HTML of the current viewport}, \texttt{user-query}, and \texttt{ground truth UI element}, respectively. 
A system, $\Psi$, has to retrieve the most important UI element as, 
\begin{equation}
    e = \Psi(w, q),
    \label{eq:data}
\end{equation}
where $\hat{e}$ is the retrieved UI element, which is compared with $e$ to compute the performance of $\Psi$.


\subsection{Dataset}
We use Mind2Web \cite{deng2023mind2web} dataset for our experiments. Mind2Web has 2,000 open-ended and real-world tasks collected from 137 websites -- making it particularly suitable for our experiment. The dataset contains three different test sets, namely-- 1) Cross-Task: examples from unseen tasks during training; 2) Cross-Website: examples from unseen websites; 3) Cross-Domain: examples from unseen domains. In our setting, these three variants of test set are especially helpful for understanding the generalizability of LLMs. We refer viewers to the original paper for more information about the dataset~\cite{deng2023mind2web}. 
Please note that in our study, we use the ``\texttt{Target Element}'' provided in the dataset as  $e$ in Equation \ref{eq:data}, since they are analogous.  
Similar to the original setting of Mind2web, we also take the rate limit of \texttt{Claude2} API into account and use a subset with 124 examples for our experiments.

\subsection{Evaluation Metric}

\begin{figure}
    \centering \includegraphics[width=0.6\textwidth,keepaspectratio]{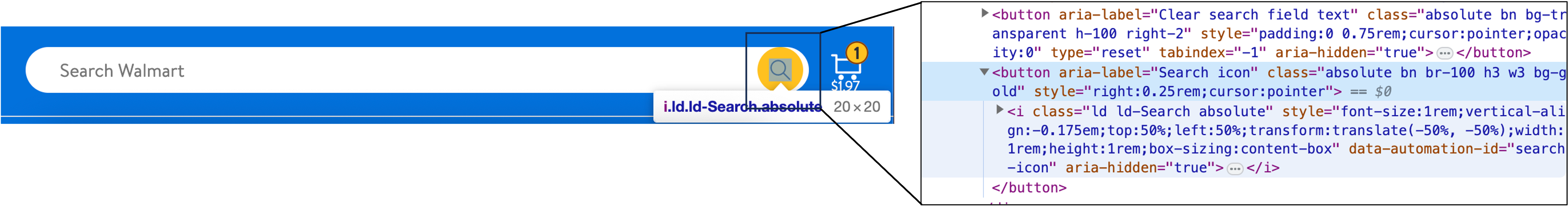}
    \caption{Example of a nested DOM layout where predicting either of parent and child can achieve identical result.}
    \label{fig:dom_example}
    \vspace{-10pt}
\end{figure}

HTML layouts are often complicated and nested in a manner where multiple elements can point to the same information. For example, Figure \ref{fig:dom_example} shows an example from \texttt{Walmart.com} where the search button has a single child element (the search icon). Predicting either the search button or the icon
would yield identical results. 
To address this, 
we take the following approach -- 
Given a UI element as the prediction, we expand the tree to get the \textit{leaf nodes} under it and compare them with the same from our ground-truth labels. 

We report the evaluation in terms of \textit{recall} (Eq.~\ref{eq:recall}) and \textit{element accuracy}. {Similar to \cite{zhou2023webarena} and \cite{Bekker_Davis_2020}, we also emphasize using \textit{recall} rather than \textit{precision} as we have access to only one positively labeled UI element.}
\begin{equation}
    \label{eq:recall}
    {\rm Recall = \frac{|Predicted\_Elements \cap Ground\_Truth|} { |Ground\_Truth|}}
\end{equation}
\textit{Element accuracy}. We follow the same approach as Mind2Web~\cite{deng2023mind2web} to calculate Element accuracy  which measures how many elements the models accuractely predicted in percentage. \footnote{Please note that the other metrics (Success Rate, Step Success Rate, Operation F1) used in Mind2Web are not well-suited to our context -- as we are not focusing on web automation.} 

\subsection{Investigation on Different Components of a Prompt}
In this section, we discuss how we probe LLM for each specific component and report the findings for each condition (denoted by \faIcon[regular]{lightbulb} icon).

    \subsubsection{\textbf{Impact of Example Selection for In-Context Learning}} 
    In-context learning (ICL) refers to a particular approach to prompt engineering in which the model is given a few examples of the task as a component of the prompt~\cite{gpt3, nakano2021webgpt}. 
    \cite{palmmt} showed that the choice of prompt examples can impact the effectiveness of ICL, and thus, the performance of LLMs. 
    If the few-shot examples in the prompt are semantically close to the target sentence, they noticed the performance improved compared to passing a set of constant examples. Inspired by their findings, we also investigate the impact of example selection in the few-shot prompting in our task. 
    To find relevant examples, we use the k-nearest neighbor search on the train set of Mind2Web, 
    Specifically, given the training examples $S$ and the target sentence $t$, we aim to find a subset $\mathcal{N}_{s,t} \subset \mathcal{S}$ with $k$ examples that are the closest to $t$, 
    %
    \begin{equation}
        \mathcal{N}^*_{s,t} = \argmin_{\mathcal{N}_{s,t} \subset \mathcal{S} : |\mathcal{N}_{s,t}| = k } \sum_{e \in \mathcal{N}_{s,t}} ~\sum_{i=1}^{|\text{Emb}(e)|} (\text{Emb}(e)_i - \text{Emb}(t)_i)^2
    \end{equation}
    where Emb$(\cdot) \in \mathbb{R}^{384}$ is the embedding space of MiniLMv2~\cite{minilmv2}, and $\mathcal{N}^*_{s,t}$ is the most optimal subset.  

    Table \ref{tab:example_accuracy} reports the performance of \texttt{Claude2} based on the choice of examples. \faIcon[regular]{lightbulb} Providing the kNN examples from the train set significantly helps in the Cross-Task test set. This might be due to the presence of similar tasks in the train set that boosts the performance \cite{palmmt}. However, for Cross-Website and Cross-Domain splits, the kNN examples did not help much. We hypothesize this might be due to the unfamiliarity of website and domain between the train and test sets. 
    
    We also noticed that for the kNN example selection, the 1-shot prompt yielded better performance than 2-shot prompt. 
    One probable reason might be 
    the increased length of input sequence causing a decay in performance, as also demonstrated by \cite{liu2023lost}. Our experiment to find the impact of HTML length on the LLM's performance support this possibility as well (see Section~\ref{sec:html_element_count} in appendix).

    However, for fixed prompt examples, the accuracy consistently improved throughout the three test sets when we used 2-shot prompting.\footnote{The few-shot examples chosen in the experiment (referred as \textit{Fixed}) are shown in Figure \ref{fig:gpt-prompt-fixed}.} We can thus understand that providing semantically relevant examples in the prompt is not enough for Web UI, we also need to be mindful of the context length.
    
\begin{table}[h]
\centering
\resizebox{0.9\textwidth}{!}{%
\begin{tabular}{lllllllllll}
\hline
\multirow{2}{*}{\textbf{Model}} &
  \multirow{2}{*}{\textbf{\begin{tabular}[c]{@{}l@{}}Prompt\\ Example\end{tabular}}} &
  \multirow{2}{*}{\textbf{\begin{tabular}[c]{@{}l@{}}Prompt \\ Setup\end{tabular}}} &
  \multicolumn{2}{c}{\textbf{Cross-Task}} &
   &
  \multicolumn{2}{l}{\textbf{Cross-Website}} &
   &
  \multicolumn{2}{l}{\textbf{Cross-Domain}} \\ \cline{4-5} \cline{7-8} \cline{10-11} 
 &
   &
   &
  \textbf{Recall} &
  \textbf{Ele. Acc} &
  \textbf{} &
  \textbf{Recall} &
  \textbf{Ele. Acc} &
   &
  \textbf{Recall} &
  \textbf{Ele. Acc} \\ \cline{1-5} \cline{7-8} \cline{10-11} 
\multirow{4}{*}{\begin{turn}{-90}\texttt{\scriptsize{Claude 2}}\end{turn}} &
  \multirow{2}{*}{Fixed} &
  1-shot &
  43.243\% &
  36.364\% &
   &
  \textit{40.909}\% &
  \textbf{40.909}\% &
   &
  11.429\% &
  17.647\% \\
 &
   &
  2-shot &
  \textbf{58.621}\% &
  \textbf{50.0}\% &
   &
  \textbf{48.235}\% &
  30\% &
   &
  \textbf{51.515}\% &
  27.273\% \\ \cline{2-11} 
 &
  \multirow{2}{*}{\begin{tabular}[c]{@{}l@{}}kNN\\  (mind2web)\end{tabular}} &
  1-shot &
  \textit{52.941}\% &
  40\% &
   &
  36.364\% &
  \textit{38.235}\% &
   &
  25.806\% &
  \textbf{29.412}\% \\
 &
   &
  2-shot &
  45.455\% &
  \textit{42.105}\% &
   &
  12.698\% &
  23.529\% &
   &
  13.725\% &
  26.667\% \\ \hline
\end{tabular}%
}
\caption{Comparison of Example selection strategy for few-shot prompting on Mind2Web Test sets.}
\label{tab:example_accuracy}
\vspace{-10pt}
\end{table}
    
\subsubsection{Impact of Specificity in User Query} 
\begin{figure}
        \centering
        \includegraphics[width=0.6\textwidth,keepaspectratio]{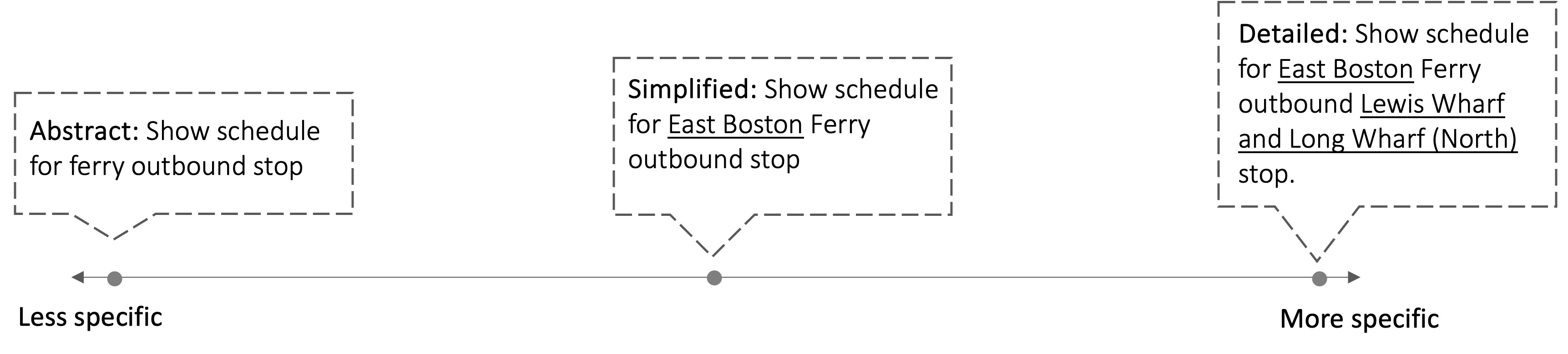}
        \caption{Example of a task description based on its level of specificity}
        \label{fig:los}
        \vspace{-10pt}
    \end{figure}

    Recent studies have shown that generic users find it difficult to prompt effectively and often opt for \textit{abstract} or \textit{higher-level} description while interacting with LLMs \cite{johnnyprompt}. However, most of the publicly available UI datasets including Mind2Web include very detailed task descriptions which can be unrealistic and unnatural. To explore the effect of specificity in LLM's performance, we augment Mind2Web task description in three ways - 1)  \textit{Detailed}: the original task description from the test set; 2) \textit{Simplified}: simplified task description which may still include essential details 3) \textit{Abstract}: High-level task description which does not include any details to mimic real-world queries by users \cite{stylette, johnnyprompt}. Figure \ref{fig:los} shows an example from the Mind2web dataset and its three variants of description. We use GPT-4 to construct the corresponding simplified and abstract descriptions. See appendix (section \ref{subsec:gpt-prompt}) for more details regarding the description creation process. To the best of our knowledge, this is the first work to investigate the impact of specificity for UI related task for LLM.

    \begin{table}[]
    \centering
    \resizebox{0.95\textwidth}{!}{%
    \begin{tabular}{llllllllllll}
    \hline
    \multirow{2}{*}{\textbf{Model}} &
      \multirow{2}{*}{\textbf{\begin{tabular}[c]{@{}l@{}}Prompt\\ Example\end{tabular}}} &
      \multirow{2}{*}{\textbf{\begin{tabular}[c]{@{}l@{}}Level of \\ Specificity\end{tabular}}} &
      \multirow{2}{*}{\textbf{\begin{tabular}[c]{@{}l@{}}Prompt \\ Setup\end{tabular}}} &
      \multicolumn{2}{c}{\textbf{Cross-Task}} &
       &
      \multicolumn{2}{l}{\textbf{Cross-Website}} &
       &
      \multicolumn{2}{l}{\textbf{Cross-Domain}} \\ \cline{5-6} \cline{8-9} \cline{11-12} 
     &
       &
       &
       &
      \textbf{Recall} &
      \textbf{Ele. Acc} &
      \textbf{} &
      \textbf{Recall} &
      \textbf{Ele. Acc} &
       &
      \textbf{Recall} &
      \textbf{Ele. Acc} \\ \cline{1-6} \cline{8-9} \cline{11-12} 
    \multirow{12}{*}{\begin{turn}{-90}\texttt{Claude2}\end{turn}} &
      \multirow{6}{*}{Fixed} &
      \multirow{2}{*}{Detailed} &
      1-shot &
      43.243\% &
      36.364\% &
       &
      40.909\% &
      \textit{40.909}\% &
       &
      11.429\% &
      17.647\% \\
     &
       &
       &
      2-shot &
      \textbf{58.621}\% &
      \textbf{50.0\%} &
       &
      \textit{48.235}\% &
      30\% &
       &
      \textbf{51.515}\% &
      27.273\% \\ \cline{3-12} 
     &
       &
      \multirow{2}{*}{Simplified} &
      1-shot &
      \textit{54.701}\% &
      38.202\% &
       &
      44.384\% &
      36.490\% &
       &
      36.087\% &
      \textit{29.474}\% \\
     &
       &
       &
      2-shot &
      53.125\% &
      43.333\% &
       &
      \textbf{49.412}\% &
      \textbf{46.341}\% &
       &
      34.211\% &
      \textbf{44}\% \\ \cline{3-12} 
     &
       &
      \multirow{2}{*}{Abstract} &
      1-shot &
      44.545\% &
      \textit{47.619}\% &
       &
      47.963\% &
      40.456\% &
       &
      \textit{39.720}\% &
      31.579\% \\
     &
       &
       &
      2-shot &
      50\% &
      38.095\% &
       &
      36.047\% &
      30.952\% &
       &
      9.523\% &
      14.286\% \\ \cline{2-12} 
     &
      \multirow{6}{*}{\begin{tabular}[c]{@{}l@{}}kNN\\ (mind2web)\end{tabular}} &
      \multirow{2}{*}{Detailed} &
      1-shot &
      \textbf{52.941}\% &
      \textit{40}\% &
       &
      36.364\% &
      \textbf{38.235}\% &
       &
      \textit{25.806}\% &
      \textbf{29.412}\% \\
     &
       &
       &
      2-shot &
      \textit{45.455}\% &
      \textbf{42.105}\% &
       &
      12.698\% &
      23.529\% &
       &
      13.725\% &
      26.667\% \\ \cline{3-12} 
     &
       &
      \multirow{2}{*}{Simplified} &
      1-shot &
      44.444\% &
      31.25\% &
       &
      15.278\% &
      30.556\% &
       &
      \textbf{30.0}\% &
      \textit{23.529}\% \\
     &
       &
       &
      2-shot &
      21.429\% &
      20.833\% &
       &
      \textbf{38.961}\% &
      35.135\% &
       &
      10.345\% &
      12.5\% \\ \cline{3-12} 
     &
       &
      \multirow{2}{*}{Abstract} &
      1-shot &
      14.286\% &
      6.667\% &
       &
      21.212\% &
      36.842\% &
       &
      6.667\% &
      13.333\% \\
     &
       &
       &
      2-shot &
      27.273\% &
      26.316\% &
       &
      19.298\% &
      32.353\% &
       &
      3.030\% &
      4.545\% \\ \hline
    \end{tabular}%
    }
    \caption{Comparison of user description specificity on Mind2Web Test sets.}
    \label{tab:specificity_accuracy}
    \vspace{-8pt}
    \end{table}

    Table \ref{tab:specificity_accuracy} reports the performance of \texttt{Claude2} based on the level of specificity in user query. \faIcon[regular]{lightbulb} For kNN based prompt example, the performance gradually decayed from detailed to abstract specificity across three test sets. However, when the examples were fixed, the result was relatively consistent throughout all levels of specificity. We hypothesize that this is due to the particular choice of the fixed examples. The task descriptions of the fixed examples are more abstract while the examples from Mind2Web train set are more detailed. This abstract task description in the fixed examples perhaps helped \texttt{Claude2} to generalize better across the three levels. This also highlights the sensitivity of LLMs towards the design of few-shot examples.
    
    \subsubsection{\textbf{Impact of HTML Truncation}} 
    \label{subsubsec:truncation}
    Raw HTML can include thousands of elements per page and such a large bulk of information can become difficult to process using LLM models \cite{deng2023mind2web}. Even in our reported accuracy from Table \ref{tab:specificity_accuracy}, we notice that 2-shot prompting hurts the performance - most likely due the the increased length of input sequence. Thus, it might be useful to filter out uninformative elements and truncate the HTML before passing it to the LLM. To test this hypothesis, we experiment with a pretrained HTML filtering module proposed by Mind2Web. This module returns the top-$k$ elements from the HTML and creates a smaller snapshot of the DOM. In our experiments, we use $k = [10, 50]$ as majority of the data samples have the ground truth elements among this region (see Section \ref{sec:filtering_mind2web} in appendix).

    \begin{table}[]
    \centering
    \resizebox{0.95\textwidth}{!}{%
    \begin{tabular}{lllllllllll}
    \hline
    \multirow{2}{*}{\textbf{Model}} &
      \multirow{2}{*}{\textbf{\begin{tabular}[c]{@{}l@{}}HTML\\ Truncation\end{tabular}}} &
      \multirow{2}{*}{\textbf{\begin{tabular}[c]{@{}l@{}}Prompt \\ Setup\end{tabular}}} &
      \multicolumn{2}{c}{\textbf{Cross-Task}} &
       &
      \multicolumn{2}{l}{\textbf{Cross-Website}} &
       &
      \multicolumn{2}{l}{\textbf{Cross-Domain}} \\ \cline{4-5} \cline{7-8} \cline{10-11} 
     &
       &
       &
      \textbf{Recall} &
      \textbf{Ele. Acc} &
      \textbf{} &
      \textbf{Recall} &
      \textbf{Ele. Acc} &
       &
      \textbf{Recall} &
      \textbf{Ele. Acc} \\ \cline{1-5} \cline{7-8} \cline{10-11} 
    \multirow{6}{*}{\begin{turn}{-90}\texttt{Claude2}\end{turn}} &
      \multirow{2}{*}{No Truncation} &
      1-shot &
      43.243\% &
      36.364\% &
       &
      40.909\% &
      40.909\% &
       &
      11.429\% &
      17.647\% \\
     &
       &
      2-shot &
      \textbf{58.621}\% &
      50.0\% &
       &
      \textit{48.235}\% &
      30\% &
       &
      \textbf{51.515}\% &
      27.273\% \\ \cline{2-11} 
     &
      \multirow{2}{*}{\begin{tabular}[c]{@{}l@{}}Truncation at \\ Top-10\end{tabular}} &
      1-shot &
      38.297\% &
      46.667\% &
       &
      34.783\% &
      37.5\% &
       &
      20\% &
      \textit{33.333}\% \\
     &
       &
      2-shot &
      43.75\% &
      \textit{51.613}\% &
       &
      34.375\% &
      38.462\% &
       &
      22.807\% &
      37.5\% \\ \cline{2-11} 
     &
      \multirow{2}{*}{\begin{tabular}[c]{@{}l@{}}Truncation at \\ Top-50\end{tabular}} &
      1-shot &
      40\% &
      43.75\% &
       &
      43.182\% &
      \textbf{50}\% &
       &
      30.556\% &
      30\% \\
     &
       &
      2-shot &
      \textit{48.649}\% &
      \textbf{51.724}\% &
       &
      \textbf{59.77}\% &
      \textit{45.238}\% &
       &
      \textit{47.368}\% &
      \textbf{43.333}\% \\ \hline
    \end{tabular}%
    }
    \caption{Comparison of Truncation strategies for few-shot prompting on Mind2Web Test sets.}
    \label{tab:truncation_accuracy}
    \vspace{-12pt}
    \end{table}

    Table \ref{tab:truncation_accuracy} reports the results for varying levels of truncation. \faIcon[regular]{lightbulb}{} Truncation significantly improved the performance compared to cases when no truncation was performed. It also resolved the issue with 2-shot prompting observed in table \ref{tab:specificity_accuracy} by effectively reducing the input token size in the LLM prompt.

     \subsubsection{ \textbf{Impact of the `Role' Assumed by the LLM}}  
     This analysis was motivated by the fact that-- 
     the perception of ``relevant UI elements'' might vary from user to user~\cite{Maioli_2018}. For example, a UI/UX designer cares more about the interaction flow of the application whereas a general user might only be interested in finding their information as soon as possible. To simulate these roles, we investigate three different personas in this paper - 1) Generic User; 2) Web Assistant; 3) UI Designer. Persona is a linguistic representation of a fictional individual, enabling LLM to mimic the actions of this imaginary character \cite{cheng2023marked}. For each persona, we generate multiple candidate prompts and empirically pick the best prompt by observing the performance on a subset of the validation set (see Section~\ref{sec:persona_explore} in appendix for details). Our final prompts for each persona are shown in Table \ref{tab:persona-table}

    \begin{table}[h]
    \centering
    \resizebox{0.9\textwidth}{!}{%
    \begin{tabular}{p{.15\linewidth}p{.85\linewidth}}
    \hline
    \textbf{Persona} & \textbf{Prompt} \\ \hline
    Web Assistant &
      You are a \hlc[cyan!20]{web assistant designed to help users navigate a website and troubleshoot issue}. Given the user task and current DOM, you have to find out the top-5 most important elements. \\ \\
    Generic User &
      You are \hlc[cyan!20]{a user surfing a website}. You have a specific goal for which you want to find out the top-5 most important elements in the current DOM. \\ \\
    UI Designer &
      You are a \hlc[cyan!20]{UI designer working to improve the user experience of a website}. Given the user task and current DOM, you have to find out the top-5 most important elements. \\ \hline
    \end{tabular}%
    }
    \caption{Prompt message for each persona.}
    \label{tab:persona-table}
    \vspace{-5pt}
    \end{table}

\begin{table}[]
\centering
\resizebox{0.95\textwidth}{!}{%
\begin{tabular}{lllllllllll}
\hline
\multirow{2}{*}{\textbf{Model}} &
  \multirow{2}{*}{\textbf{Persona}} &
  \multirow{2}{*}{\textbf{\begin{tabular}[c]{@{}l@{}}Prompt \\ Setup\end{tabular}}} &
  \multicolumn{2}{c}{\textbf{Cross-Task}} &
   &
  \multicolumn{2}{l}{\textbf{Cross-Website}} &
   &
  \multicolumn{2}{l}{\textbf{Cross-Domain}} \\ \cline{4-5} \cline{7-8} \cline{10-11} 
 &
   &
   &
  \textbf{Recall} &
  \textbf{Ele. Acc} &
  \textbf{} &
  \textbf{Recall} &
  \textbf{Ele. Acc} &
   &
  \textbf{Recall} &
  \textbf{Ele. Acc} \\ \cline{1-5} \cline{7-8} \cline{10-11} 
\multirow{6}{*}{\begin{turn}{-90}\texttt{Claude2}\end{turn}} &
  \multirow{2}{*}{Web Assistant} &
  1-shot &
  35.135\% &
  33.333\% &
   &
  \textbf{61.25}\% &
  \textbf{60}\% &
   &
  15.625\% &
  24.138\% \\
 &
   &
  2-shot &
  40.741\% &
  \textit{43.478}\% &
   &
  29.069\% &
  41.463\% &
   &
  \textbf{60.417}\% &
  \textit{25}\% \\ \cline{2-11} 
 &
  \multirow{2}{*}{Generic User} &
  1-shot &
  \textit{41.935}\% &
  39.286\% &
   &
  37.349\% &
  25.641\% &
   &
  7.813\% &
  13.793\% \\
 &
   &
  2-shot &
  \textbf{73.077}\% &
  \textbf{55.556}\% &
   &
  \textit{48.529}\% &
  40\% &
   &
  13.235\% &
  22.581\% \\ \cline{2-11} 
 &
  \multirow{2}{*}{UI Designer} &
  1-shot &
  35.714\% &
  30.769\% &
   &
  15.294\% &
  31.707\% &
   &
  12.069\% &
  20\% \\
 &
   &
  2-shot &
  32.143\% &
  32\% &
   &
  42.857\% &
  \textit{45}\% &
   &
  \textit{30.159}\% &
  \textbf{29.630}\% \\ \hline
\end{tabular}%
}
\caption{Comparison of choice of persona on Mind2Web Test sets.}
\label{tab:persona_accuracy}
\vspace{-12pt}
\end{table}

    Table \ref{tab:persona_accuracy} reports the performance of \texttt{Claude2} based on the choice of persona in input prompt. \faIcon[regular]{lightbulb} Web Assistant persona performed significantly better (except for the User Persona for 2-shot prompting on Cross-Task) than the rest of the persona. 


\subsubsection{Error Patterns in LLM Response}

\begin{figure}[h]
             \centering
             \begin{subfigure}[b]{0.45\textwidth}
                 \centering
                 \includegraphics[width=0.85\textwidth,keepaspectratio]{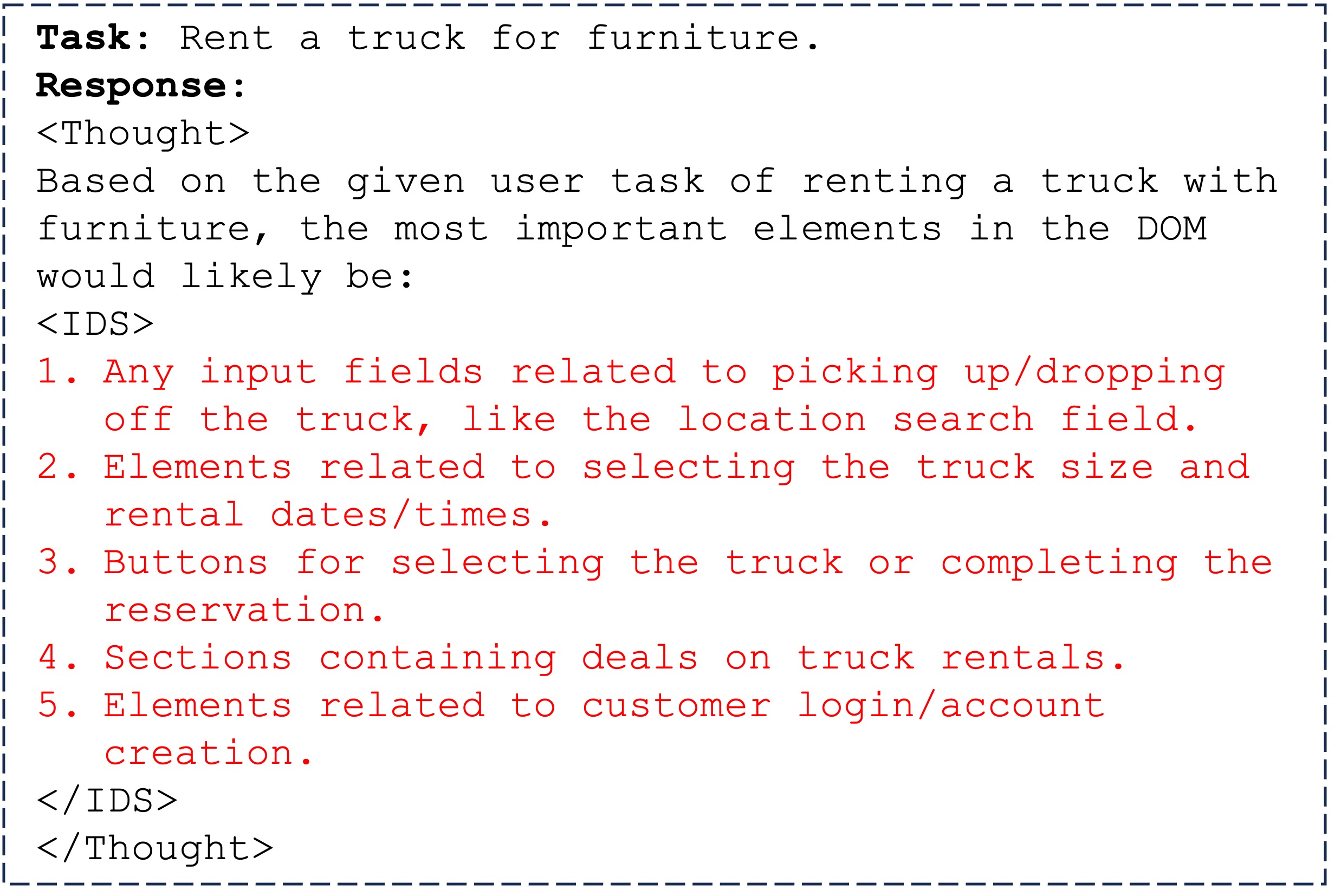}
                 \caption{LLM gave proper reasoning, but didn't give the ID of the HTML elements as instructed.}
             \label{subfig:a}
             \end{subfigure}
             \hfill
             \begin{subfigure}[b]{0.45\textwidth}
                 \centering
                 \includegraphics[width=0.75\textwidth,keepaspectratio]{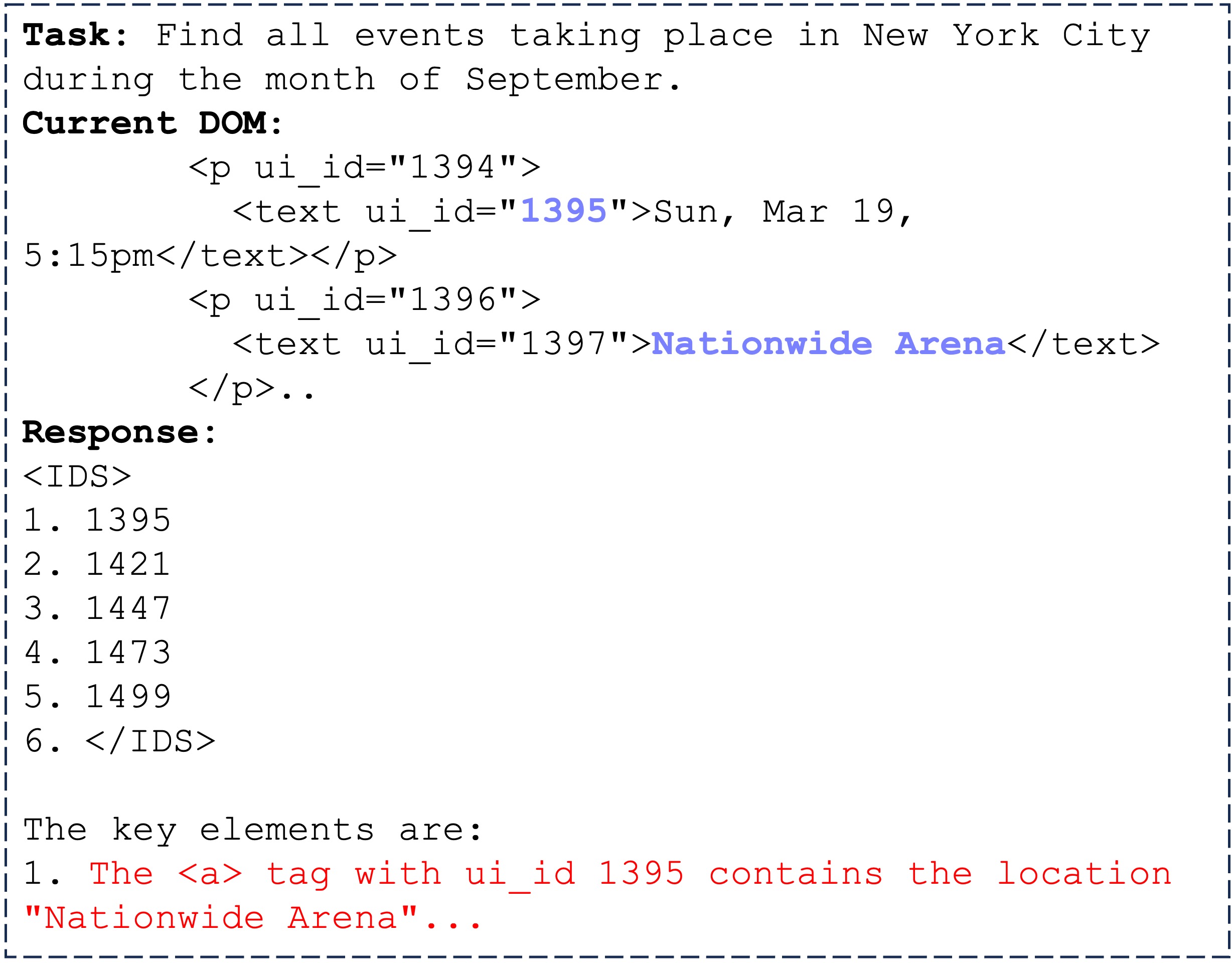}
                 \caption{Reference to wrong DOM element that was hallucinated from other elements in the DOM.)}
                 \label{subfig:b}
             \end{subfigure}
             
                \caption{Examples of frequently occurring errors. {We show only the relevant fractions of the outputs for the ease of the readers.}}
                \label{fig:examples_failure}
        \vspace{-10pt}
        \end{figure}

During our experiments, we noticed two types of errors most frequently - 1) Failure to follow instruction; and 2) Hallucination. Here we provide two such examples in Figure \ref{fig:examples_failure}. 

In the first example (Figure \ref{subfig:a}), the model failed to follow the instruction to respond with the IDs within the \texttt{<ID>} tag, rather it provided the reasoning of its choice. In the second example (Figure \ref{subfig:b}), the model hallucinated a non-existing DOM element by merging the information ({\em i.e}, nodeID and text description) from two existing elements. 

\section{Conclusions and Future Work}

In this study, we investigate the capacities of LLM for retrieving important information from a webpage given user task description. Our current study is solely based on Anthropic's \texttt{Claude2} model. In the future, we would like to extend our study to other large language models such as GPT-4\cite{openai2023gpt4}, Llama V2\cite{touvron2023llama}, Vicuna\cite{vicuna}, PaLM~2\cite{anil2023palm} and so on. One of the primary challenges in extending our study to these models lies in fitting a substantial HTML context within their comparatively limited context lengths. Future work can explore how to efficiently encode this information while preserving the syntactical and contextual integrity of the DOM. The truncation strategy we used in Section \ref{subsubsec:truncation} can be further modified for this purpose. 

A major finding of our work is the co-relation of the specificity of user query and the performance. Future work can learn from our experiment and work on making these models more responsive. i.e: knowing when to intervene or ask for further clarification. A very interesting direction we are interested in is - even if the prompts are abstract, how can we make LLMs understand the user intention better? This could entail refining the models' ability to interpret and respond to nuanced user intentions. The model can also benefit from personalization and contextualized information about the user and/or other metadata such as current location, time, or data. However, future research should also be mindful of the security concerns when using the personal information of the users.




\bibliographystyle{unsrt}
\bibliography{main}
\medskip



\appendix 

\section{Appendix}

\subsection{Design of Fixed Examples Used in Few-shot Prompting}
\label{sec:design_of_fixed_example}
Figure \ref{fig:gpt-prompt-fixed} shows the fixed examples we used in our experiment. To craft these examples, we preserved the bare-minimum information in the DOM and replaced everything else with `\texttt{..}'. 

\begin{figure}[H]
\fbox{%
    \parbox{\textwidth}{%
    \texttt{Here are two examples -\\
        User Task: fill out a registration form \\
        Current DOM: <html>..<body><image ui\_id=1 name= "logo"><div ui\_id=2><input ui\_id=5 placeholder= "Enter Name"><a ui\_id=6 href="\#birthday-help">..<select ui\_id=7 aria-label= "Month"><button ui\_id=10 name= "submit"></div>..</body>..</html>\\
        Your Response:\\
        <Thought>The most important elements for this task would be input fields (to put in the details), submission buttons (to submit the form), and help instructions (to read instructions)</Thought>\\
        <IDS>\\
        1. 2\\
        2. 5\\
        3. 7\\
        4. 10\\
        5. 6\\
        </IDS>\\
        User Task: place an order\\
        Current DOM: <html>..<body><image ui\_id=3 name= "Welcome to our website!"><div ui\_id=8><input ui\_id=5 placeholder= "Enter Product Name"><button ui\_id=10 name= "Search"></div>..<a ui\_id=25 href="\#new deals"><div ui\_id=29 href="\#sales"><span ui\_id=25>"Account Setting"</span>...</body>..</html>\\
        Your Response:\\
        <Thought>The most important elements for next step might be to fill up the search field and search button, and relevant deals.</Thought>    \\
        <IDS>\\
        1. 5\\
        2. 10\\
        3. 8\\
        4. 25\\
        5. 29\\
        </IDS>\\
        You must always include the <Thought> and <IDs> open/close tags or else your response will be marked as invalid. <IDs> must include the ui\_id number of the elements.
    }}%
}
    \caption{Fixed examples used in the input prompt.}
    \label{fig:gpt-prompt-fixed}
\end{figure}

\subsection{Exploration of Persona Prompt}
\label{sec:persona_explore}

All scores are reported on a small validation set from Cross-Task comprising on 2 tasks and 14 steps. These tasks are chosen randomly and kept constant for all the variants of prompt shown below. The performance scores on this subset, along with the candidate prompts, are shown in Tables \ref{tab:persona_webassist}, \ref{tab:persona_genericuser}, and \ref{tab:persona_uidesigner}.

\begin{table}[H]
    \centering
    \resizebox{0.9\textwidth}{!}{%
    \begin{tabular}{p{.8\linewidth}p{.15\linewidth}}
    \hline
    \textbf{Prompt} & \textbf{Recall} \\ \hline

      You are a \hlc[cyan!20]{web assistant designed to improve the user experience of a website}. Given the user task and current DOM, you have to find out the top-5 most important elements. & 33.333\% \\ \\
    
      You are a \hlc[cyan!20]{web assistant designed to help users navigate a website}. Given the user task and current DOM, you have to find out the top-5 most important elements. & 27.778\% \\ \\
    
      You are a \hlc[cyan!20]{web assistant designed to help users navigate a website and troubleshoot issue}. Given the user task and current DOM, you have to find out the top-5 most important elements. & \textbf{50.0\%}  \\ \\
      
      You are a \hlc[cyan!20]{web assistant designed to help users find relevant information}. Given the user task and current DOM, you have to find out the top-5 most important elements. & 44.444\%  \\ \hline

    \end{tabular}%
    }
    \caption{Initial Candidate prompts for Web Assistant persona.}
    \label{tab:persona_webassist}
    \end{table}

\begin{table}[H]
    \centering
    \resizebox{0.9\textwidth}{!}{%
    \begin{tabular}{p{.8\linewidth}p{.15\linewidth}}
    \hline
    \textbf{Prompt} & \textbf{Recall} \\ \hline

      You are a \hlc[cyan!20]{user surfing a website. You have a specific goal for which} you want to find out the top-5 most important elements in the current DOM. & \textbf{38.889\%}  \\ \\
      
      You are a \hlc[cyan!20]{user who is browsing a website to perform a task}. You want to find out the top-5 most important elements in the current DOM for your task. & 22.222\%  \\ \hline

    \end{tabular}%
    }
    \caption{Initial Candidate prompts for Generic User persona.}
    \label{tab:persona_genericuser}
    \end{table}
 
\begin{table}[H]
    \centering
    \resizebox{0.9\textwidth}{!}{%
    \begin{tabular}{p{.8\linewidth}p{.15\linewidth}}
    \hline
    \textbf{Prompt} & \textbf{Recall} \\ \hline

      You are a \hlc[cyan!20]{UI designer working to improve the user experience of a website}. Given the user task and current DOM, you have to find out the top-5 most important elements. & \textbf{50.0\%} \\ \\
    
      You are a \hlc[cyan!20]{UI designer working on Quality Assurance of a website}. Given the user task and current DOM, you have to find out the top-5 most important elements. & 33.333\% \\ \\
    
      You are a \hlc[cyan!20]{UI designer who wants to make user-friendly interface}. Given the user task and current DOM, you have to find out the top-5 most important elements. & 22.222\%  \\ \\
      
      You are a \hlc[cyan!20]{UI designer who wants to ensure user can find task-specific information easily. So,} for a given task and current DOM, you want to find out the top-5 most important elements. & 27.778\%  \\ \hline

    \end{tabular}%
    }
    \caption{Initial Candidate prompts for UI Designer persona.}
    \label{tab:persona_uidesigner}
    \end{table}

\subsection{Creation of Task Description Based on Varying Specificity}
\label{subsec:gpt-prompt}
We use GPT-4 to create \textit{simplified} and \textit{abstract} task description for mind2web. We use a simple few-shot prompting technique to show how to simplify the task description while preserving the meaning. The first author did a manual investigation over a generated subset of task description to ensure its quality. The specific prompt used for the creation of task description is shown below in Figure \ref{fig:gpt-prompt}. In the prompt, A1 and A2 denote \textit{simplified} and \textit{abstract} task description respectively.

\begin{figure}[H]
\fbox{%
    \parbox{\textwidth}{%
    Your task is to remove variable names in a task description. Variables can be place, date, product name and so on. Respond with the variables you removed and the simplified task description after removal. Provide two simplifications - 1) where no variable exists, 2) where you may keep some variables that are necessary to preserve the meaning.  \\
    
    Here are two examples: \\
Task description - Find the nearest Chinese restaurants in Washington DC. \\
Response - \\
<Reasoning>Chinese is a specific cuisine. Washington DC is a place. So, I need to remove them.</Reasoning> \\
<Variables>var1:  Chinese, var2: Washington DC</Variables> \\
<Template>Find the nearest [var1] restaurants in [var2] </Template>\\
<A1>Find the nearest restaurants</A1>\\
<A2>Find the nearest restaurants in Washington DC</A2>\\

Task description - Check for pickup restaurant available in Boston, NY on March 18, 5pm with just one guest.\\
Response-\\
<Reasoning>Boston, NY is a place. March 18 is a date. 5pm is a time. one is a number. So, I need to remove them.</Reasoning>\\
<Variables>var1:  Boston, NY, var2: March 18, var3: 5pm, var4: one</Variables>\\
<Template>Check for pickup restaurant available in [var1] on [var2], [var3] with just [var4] guest. </Template>\\
<A1>Check for pickup restaurant</A1>\\
<A2>Check for pickup restaurant in Boston, NY at 5pm</A2>\\

Always think step-by-step and include <Reasoning>, <Variables>, <Template>, <A1>, <A2> tags.
    }%
}
    \caption{Prompt for GPT-4 model to construct abstract and simplified task description.}
    \label{fig:gpt-prompt}
\end{figure}

\subsection{Filtering Module from Mind2Web}
\label{sec:filtering_mind2web}

\begin{figure}[H]
    \centering
    \includegraphics[width=0.6\textwidth,keepaspectratio]{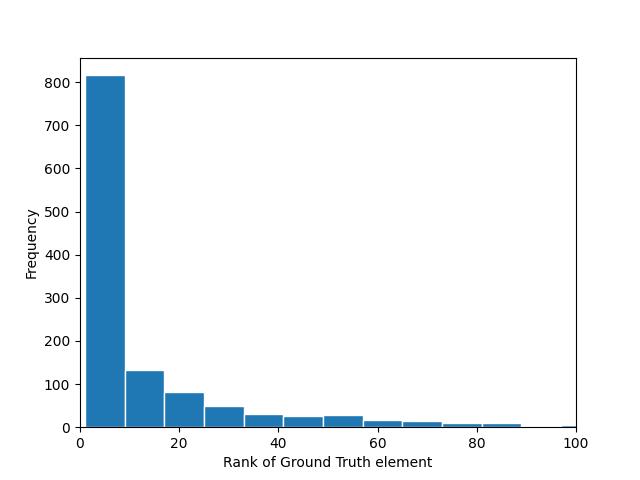}
    \caption{Distribution of Rank of ground truth elements in the prediction of Ranking Module. For majority of the data samples, the ground truth is present in within Top-10.}
    \label{fig:dist_table}
\end{figure}

The filtering module in Mind2Web paper is based on DeBERTa \cite{he2021deberta} model. We used the pretrained module provided in their paper in our analysis. We start by analyzing their accuracy as shown as Figure \ref{fig:dist_table}. For majority of the test samples, the ground truth element is present within the Top-10 elements returned by the model. Hence, we truncate the results from DeBERTa at Top-10. To understand the impact of a larger window, we also show results from Top-50 as well.

\subsection{Distribution of the HTML Element Count and Its Impact on Performance}
\label{sec:html_element_count}

For this analysis, we categorized the results into five clusters based on the number of HTML elements in each sample. Then, we calculated the sample count within each cluster along with the corresponding element accuracy

\begin{figure}[h]
    \centering
    \includegraphics[width=0.7\textwidth,keepaspectratio]{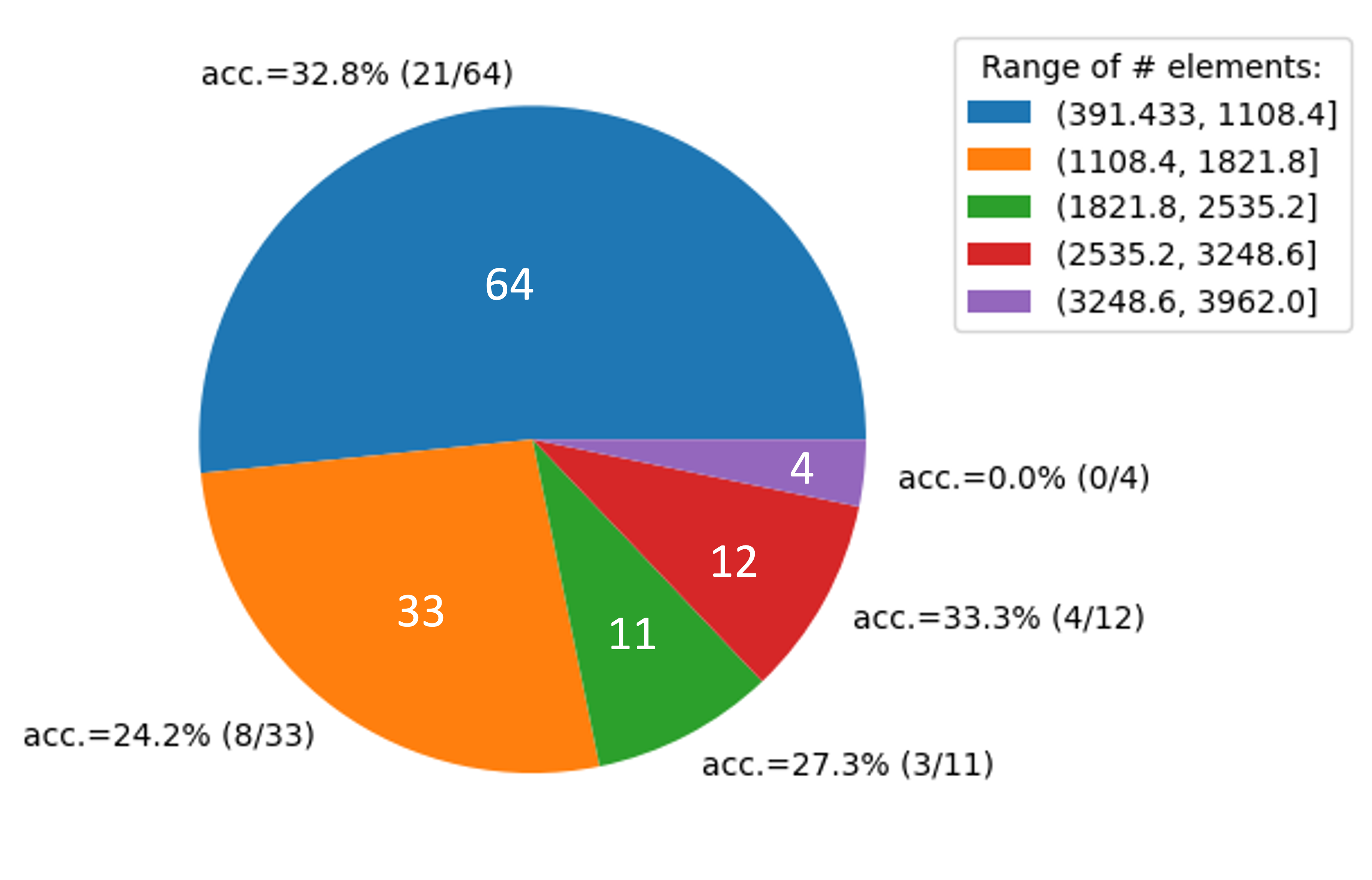}
    \caption{Distribution of Accuracy with respect to Number of HTML elements}
    \label{fig:pie_chart_dist}
\end{figure}

In Figure~\ref{fig:pie_chart_dist}, we demonstrate this analysis as a pie-chart, where each pie represents one of these five clusters. Within each pie, we show the number of samples that fall within the corresponding cluster; and the accuracy is demonstrated right outside the pie (shown as ``acc.''). The legend in the figure shows the range of the number of HTML element for each cluster. 

It is worth noting that as the number of HTML elements increases, the available samples diminish. Furthermore, it appears that the LLM's performance falters when dealing with substantially longer HTML content, specifically within the range of $(3248.6, 3962.0]$.



\end{document}